\documentclass[conference,10pt]{IEEEtran}

\usepackage{cite}
\usepackage{amsmath}
\usepackage{amssymb}
\usepackage{amsfonts}
\usepackage{amsthm}
\usepackage{graphicx}
\usepackage{cleveref}
\usepackage{nicematrix}

\theoremstyle{definition}
\newtheorem{definition}{Definition}

\newcommand{\subheading}[1]{\noindent\textbf{#1}}
\newcommand{\CCX}{\mathcal{X}}
\DeclareMathOperator{\rk}{rk}
\newcommand{\Znon}{\mathbb{Z}_{\ge 0}}

\begin{document}

\title{Combinatorial Complexes: Bridging the Gap Between Cell Complexes and Hypergraphs}


\author{
\IEEEauthorblockN{Mustafa Hajij}
\IEEEauthorblockA{University of San Francisco\\
San Francisco, USA}\\
\IEEEauthorblockN{Aldo Guzmán-Sáenz}
\IEEEauthorblockA{IBM Corporation\\
New York, USA}
\and
\IEEEauthorblockN{Ghada Zamzmi}
\IEEEauthorblockA{University of South Florida \\
Florida, USA}\\
\IEEEauthorblockN{Tolga Birdal}
\IEEEauthorblockA{Imperial College London\\
London, UK}
\and
\IEEEauthorblockN{Theodore Papamarkou}
\IEEEauthorblockA{The University of Manchester\\
Manchester, UK}\\
\IEEEauthorblockN{Michael T. Schaub}
\IEEEauthorblockA{RWTH Aachen University\\
Aachen, Germany\\
schaub@cs.rwth-aachen.de}} 

\maketitle

\begin{abstract}
Graph-based signal processing techniques have become essential for handling data in non-Euclidean spaces. However, there is a growing awareness that these graph models might need to be expanded into `higher-order' domains to effectively represent the complex relations found in high-dimensional data.
Such higher-order domains are typically modeled either as hypergraphs, or as simplicial, cubical or other cell complexes.
In this context, cell complexes are often seen as a subclass of hypergraphs with additional algebraic structure that can be exploited, e.g., to develop a spectral theory. 
In this article, we promote an alternative perspective. We argue that hypergraphs and cell complexes emphasize \emph{different} types of relations, which may have different utility depending on the application context.
Whereas hypergraphs are effective in modeling set-type, multi-body relations between entities, cell complexes provide an effective means to model hierarchical, interior-to-boundary type relations.
We discuss the relative advantages of these two choices and elaborate on the previously introduced concept of a combinatorial complex that enables co-existing set-type and hierarchical relations. 
Finally, we provide a brief numerical experiment to demonstrate that this modelling flexibility can be advantageous in learning tasks.
\end{abstract}


\vspace{-3mm}
\section{Introduction}
Graph-based signal processing and learning methods have seen a surge of interest recently~\cite{ortega2018graph,dong2020graph}. 
This interest can be partially attributed to at least two factors. 
First, graphs provide a natural abstraction for a variety of complex systems, which consist of entities (abstracted as vertices) that are coupled to each other by some form of relation or interaction (abstracted as edges). 
Second, graphs may be seen as discretizations of (non-Euclidean) spaces, such as manifolds. 
Graphs thus provide a language to articulate a vast number of problems arising in the analysis of complex systems and data.

However, it has been realized in the literature \cite{battiston2020networks,bick2021higher,torres2021and,hajijtopological,schaub2021signal} that graph-based models may not be enough to faithfully represent the rich set of relations found in complex systems and high-dimensional data in certain scenarios.
Specifically, graphs inherently describe pairwise relations (edges) between entities (vertices).
This can be limiting if there are interactions between multiple entities simultaneously or if there are couplings between the relations themselves.
Accordingly, there has been an interest in developing `higher-order' models beyond graphs that are capable of modeling these interactions more explicitly. 

Two types of models that have been predominantly considered so far are hypergraphs and simplicial/cell complexes~\cite{antelmi2023survey,bick2021higher,ebli2020simplicial,schaub2021signal,torres2021and,hajijtopological,battiston2020networks,hajijcell}.
Both types of abstraction have been used in a variety of contexts, ranging from network dynamical systems to machine learning and signal processing tasks~~\cite{antelmi2023survey,bick2021higher,battiston2020networks,ebli2020simplicial,schaub2021signal,torres2021and,hajijtopological}. 
In general, it has been found that a variety of (new) phenomena and effects can arise when higher-order relations are taken into account~\cite{julkunen2020leveraging}.
Complexes are in this context often seen as a specific type of hypergraph that has to fulfill additional closure conditions, in the sense that the presence of a higher-order relation implies the existence of all corresponding lower-order relations \cite{torres2021and}.
The upshot of this slightly more restrictive structure is that cell complexes can be endowed with spectral theory, grounded in notions from algebraic topology.

\subheading{Contributions.}
In this work, we challenge this current view that complexes are to be seen as a special form of hypergraph.
We emphasize a fresh look at the relative advantages of hypergraphs on one side---which aptly model set-type, multi-body relations among nodes---and complexes on the other side---which aptly model hierarchical interior-to-boundary-type relations. 
We discuss the relative merits of these two options and highlight their difference in terms of their associated multi-partite representations.
We also elaborate on the concept of a combinatorial complex, as introduced in~\cite{hajijtopological}, a novel type of complex that accommodates both set-type and hierarchical relations. 
To underscore the utility of combinatorial complexes, we present a numerical experiment demonstrating their advantages in a learning task.

\subheading{Related work.} 
Simplicial complexes and hypergraphs are different forms of high-order data structures~\cite{bianconi2021higher}, which have so far been the most studied higher-order extensions of graphs~\cite{antelmi2023survey,bick2021higher,battiston2020networks,ebli2020simplicial,schaub2021signal,torres2021and,hajijtopological}.
While cell complexes are typically defined via topological notions, many attempts have been made in the geometric topology literature to make the definition purely combinatorial~\cite{aschbacher1996combinatorial, basak2010combinatorial, savoy2022combinatorial}, to make them computationally more useful. 
Combinatorial complexes~\cite{hajijtopological} are inspired by geometric topology notions and relax the definition of cell complex to include set-type and hierarchical relations.

\section{Modeling higher-order relations}

\subsection{Preliminaries}
Before discussing the modeling of higher-order domains, we recap some notions from graph theory.
In the following, we are concerned with systems comprising a large set of interrelated entities,
which we model as vertices (or nodes) connected by edges.
A canonical way to represent such systems is in term of a graph $\mathcal{G}=(\mathcal{V},\mathcal{E})$ consisting of a vertex set $\mathcal{V}$ and an edge set $\mathcal{E}$. For simplicity, we identify the vertex set with the first $n$ integers $\mathcal{V} = \{1,\dots,n\}$, and consider undirected graphs, that is, $\mathcal{E} \subseteq \{\{i,j\} \mid i,j\in V\}$.

\begin{figure}[tb!]
    \centering
    \includegraphics{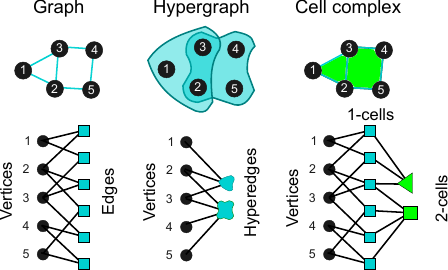}
    \caption{\textbf{Abstractions of relational data via graphs, hypergraphs and cell complexes.} \textbf{Left:} A graph may be presented via a bipartite structure comprising a set of vertices and a set of edges. Each edge is associated with exactly two nodes.
    \textbf{Middle:} A hypergraph can also be represented by a bipartite structure with a set of vertices and a set of (hyper)edges. In contrast to the case of graphs, the hyperedges do not have a cardinality constraint, i.e., we relax the constraints on the connectivity of each (hyper)edge in the bipartite representation.
    \textbf{Right:} a cell complex can in general \emph{not} be presented via a bipartite structure, but represents a multi-partite structure, in which all but the out-most layers are connected both "downwards" and "upwards" via boundary-to-interior type relations (this is simply the Hasse diagram of the complex).
    Note in particular that, e.g. a 2-simplex (triangle (1,2,3) above) \textit{does not correspond to a direct relation between three nodes}, rather it is a relation between three edges (1-cells). 
    In general, there is only an indirect coupling between cells of order $k$ and $k'$ if $|k-k'|>1$.
    Similarly to the case of graphs, there are certain cardinality constraints with respect to the connectivity between the layers (e.g., a 1-simplex always just links two nodes).\vspace{-3mm}}
    \label{fig:data-structures}
\end{figure}

Although this point of view is rarely emphasized in standard network modeling, let us point out that \emph{any} graph (not just bipartite graphs) may be encoded via a bipartite structure. 
The bipartite structure consists of a set of vertices, on the one side, and a set of pairwise relations (edges), on the other side, which couple exactly two vertices (see~\Cref{fig:data-structures}). 
Denoting by $\mathbf{B_1}$ the node-to-edge incidence matrix (either signed or unsigned), this bipartite graph can be encoded via a bipartite adjacency matrix of the form
\begin{equation}
    \mathbf{D}_\mathcal{G} = 
\begin{bNiceMatrix}[first-row,first-col]
 & \mathcal V & \mathcal E \\
\mathcal V & \mathbf{0} & \mathbf{B_1} \\
\mathcal E & \mathbf{B_1}^\top & \mathbf{0} \\
\end{bNiceMatrix} \in \mathbb{R}^{(|\mathcal V|+|\mathcal E|)\times (|\mathcal V|+|\mathcal E|)}.
\end{equation}
This somewhat unusual viewpoint is instructive to highlight the relative differences in extending graphs towards hypergraphs or complexes, as we discuss next.

\subsection{Hypergraphs: set-type relations between multiple nodes}
Unlike graphs, which only encode pairwise relations between vertices, hypergraphs can describe multi-vertex relations.

\begin{definition}[Hypergraph]
\label{hyperG:main}
A \textit{hypergraph} $\mathcal{H}$ on a nonempty set $\mathcal{V}$ is a pair $(\mathcal{V}, \mathcal{E}_\mathcal{H}$), where $\mathcal{E}_\mathcal{H}$ is a non-empty subset of the powerset $\mathcal{P}(\mathcal{V})\setminus\{\emptyset\}$ of the vertices. Elements of $\mathcal{E}_\mathcal{H}$ are called \textit{hyperedges}.
\end{definition}

We can see from Definition \ref{hyperG:main} that a relation in a hypergraph can involve any subset of vertices, not necessarily following a predefined pattern.
Hypergraphs thus offer a way to represent relations between vertices, essentially without any imposed structure~\cite{Klamt2009}.
This flexibility is of high utility for modeling many types of co-dependencies between variables associated to nodes.
For example, (probabilistic) factor graphs~\cite{loeliger2004introduction} may be seen as an instance of a hypergraph structure: indeed, the bipartite representation of the hypergraph shown in~\Cref{fig:data-structures} is precisely of the form of a factor graph, where the hyperedges play the role of factors.
A closely related example are graph-based codes~\cite{kschischang2003codes,loeliger2004introduction}, in which message nodes (vertices) are related to check nodes (hyperedges) which describe parity bits (which are a function of all associated message nodes).
Similarly, hypergraph-based formalism has been considered extensively when modeling distributed (network) dynamics, involving multiple entities (see, e.g.,~\cite{boccaletti2023structure}).
A well studied example of such dynamics, in which the hypergraph structure is, however, not always made explicit, is chemical reaction networks~\cite{muller2022makes,Klamt2009}.

The bipartite view on hypergraphs~\cite{Klamt2009}, displayed in~\Cref{fig:data-structures}, is encoded in a corresponding adjacency matrix as
\begin{equation}
    \mathbf{D}_{\mathcal{G}_\mathcal{H}} = 
\begin{bNiceMatrix}[first-row,first-col]
 & \mathcal V & \mathcal E_\mathcal{H} \\
\mathcal V & \mathbf{0} & \mathbf{B}_\mathcal{H} \\
\mathcal E_\mathcal{H} & \mathbf{B}_\mathcal{H}^\top & \mathbf{0} \\
\end{bNiceMatrix} \in \mathbb{R}^{(|\mathcal V|+|\mathcal{E}_\mathcal{H}|)\times (|\mathcal V|+|\mathcal{E}_\mathcal{H}|)},
\end{equation}
where $\mathbf{B}_\mathcal{H}$ is the corresponding vertex-to-hyperedge incidence matrix, that encodes which vertices belong to which hyperedge.
This perspective also highlights the duality between vertices and hyperedges, in that we may think of the nominal vertices as our entities whose relations are defined via the hyperedges, or vice versa, we may consider the nominal hyperedges as our entities, which are linked to each other via the nominal vertices. 
Indeed, this type of duality underlies a number of `line graph'-like transformations in which the role of vertices and hyperedges is switched (see, e.g.,~\cite{schaub2021signal} for further discussion on such transformations).

However, the large flexibility of hypergraphs also implies that there is a large variety of algebraic representations that can be associated with them, including a large number of Laplacian matrices~\cite{banerjee2023}, but also tensor respresentations (cf. \cite{schaub2021signal}). 
As a consequence, spectral theory for hypergraphs is still comparably less well developed.
Unlike graphs, which may be readily interpreted as a discretization or approximation of an underlying (latent) continuous space, a geometric interpretation of hypergraphs is more elusive, which contributes also to the difficulty of developing a coherent notion of a Laplacian operator, which is intimately related to geometric and topological ideas.
This contrasts with cell complexes, which have a strong grounding in geometry and (algebraic) topology, as we will discuss next.



\subsection{Cell complexes: hierarchical interior-to-boundary relations}

Different to hypergraphs, cell complexes have a clear topological and geometric grounding~\cite{hatcher2005algebraic}: at each level, they couple elements of a boundary to an interior. 
Arguably, the simplest variant of a cell complex is a simplicial complex~(SC).

\begin{definition}[Abstract simplicial complex]
\label{SCs:main}
An \textit{abstract simplicial complex (SC)}
in a non-empty set $S$ is a pair $SC=(S,\mathcal{X})$, where $\mathcal{X}$ is a subset of $\mathcal{P}(S) \setminus \{\emptyset\}$ such that $ x \in SC$ and $y \subseteq x $ imply $y \in SC$. 
Elements of $\mathcal{X}$ are called \textit{simplices}.
\end{definition}

Simplicial complexes~\cite{boissonnat2018geometric} enable the formation of hierarchical structures, as they can include relations that encompass other relations. 
The hierarchical aspects of a SC are apparent from the closure property with respect to subsets of the SC ($x\in SC, y\subseteq x \implies y\in SC$).
This hierarchical structure is important for numerous applications, as it offers a method for organising and interpreting data.
Though commonly portrayed otherwise, it is important to clarify that, from the perspective of boundary-to-interior relations, a 2-simplex is generally \emph{not} to be interpreted as a relation between three nodes. 
Instead, it should be seen as a relation among the three edges that constitute the boundary of the 2-simplex, where the endpoints of these edges correspond to the three nodes.
This becomes apparent, when considering the multi-partite representation of a SC in terms of its boundary maps (cf.~\Cref{fig:data-structures} for a cell complex), which always link $k$-simplices to $k+1$ simplices; i.e., there is no direct (boundary) relation between $0$-simplices and general $k$-simplexes for $k\neq 1$.

Simplicial complexes carry cardinality constraints for its relations in that a $k$-simplex $\varsigma_k$ always contains $k+1$ simplices of order $k-1$. 
In fact, since $k+1$ is exactly the number of $0$-simplices $\varsigma_k$ contains, there is a one-to-one mapping between the set of $0$-simplices in $\varsigma_k$ and the corresponding set of $k-1$ simplices defining the boundary of the $k$-simplex $\varsigma_k$. 
While this makes for a convenient combinatorial definition of SCs, and SCs are sufficient to discretize essentially all important topological spaces, the restriction to simplices is an arbitrary condition.
Cell complexes remove this restriction and thus enable a more flexible and compact representation of a (hierarchical) set of relations.

Regular cell complexes extend the idea of simplicial complexes by providing more flexibility in the shapes of relations. Instead of being limited to `triangles' and their higher-dimensional combinatorial counterparts, we can use various types of shapes, making it a more versatile representation.

\begin{definition}[Regular cell complex]
\label{RCC:main}
A \textit{regular cell complex} is a topological space $S$ with a partition into subspaces (\textit{cells}) $\{x_\alpha\}_{\alpha \in P_{S} }$, where $P_{S}$ is an index set, satisfying the following conditions:
\begin{enumerate}
\item $S= \cup_{\alpha \in P_{S}} \text{int}(x_{\alpha})$,
where $int(x)$ denotes the interior of cell $x$.
\item For each $\alpha \in P_S$,
there exists a homeomorphism $\psi_{\alpha}$, called an \textit{attaching map}, 
from $x_\alpha$ to $\mathbb{R}^{n_\alpha}$ for some $n_\alpha\in \mathbb{N}$,
called the \textit{dimension} $n_\alpha$ of cell $x_\alpha$.
\item For each cell $x_{\alpha}$,
the boundary $\partial x_{\alpha}$ is a union of finitely many cells,
each having dimension less than the dimension of $x_{\alpha}$. 
\end{enumerate}
\end{definition}

The multi-partite structure of a cell complex can be easily visualised as a multi-partite graph (see~\Cref{fig:data-structures}, right), isomorphic to the Hasse diagram of the complex.
In terms of an adjacency matrix, this can be written as
\begin{equation}\label{eq:dirac_cell}
    \mathbf{D}_\mathcal{X} = 
\begin{bNiceMatrix}[first-row,first-col]
 & \mathcal V & \mathcal X_1 & \mathcal X_2 \\
\mathcal V & \mathbf{0} & \mathbf{B_1} & \mathbf{0} \\
\mathcal X_1 & \mathbf{B_1}^\top & \mathbf{0} & \mathbf{B_2} \\
\mathcal X_2 & \mathbf{0} & \mathbf{B_2}^\top & \mathbf{0} \\
\end{bNiceMatrix} \in \mathbb{R}^{N_\mathcal{X}\times N_\mathcal{X}},
\end{equation}
where $\mathcal{X}_i$ denotes the set of cells of order $i$, $N_\mathcal{X} = |\mathcal V|+|\mathcal X_1| + |\mathcal X_2|$ denotes the number of all cells contained (including the vertices) in the complex, and $\mathbf{B_i}$ is the $i$th boundary matrix mapping cells of order $i$ to cells of order $i-1$.
Note that these boundary matrices are typically equipped with an orientation and meet the condition $\mathbf{B_1}\mathbf{B_2}=0$ (the boundary of a boundary is empty)~\cite{hatcher2005algebraic,grady2010discrete}.
We remark that the matrix in \cref{eq:dirac_cell} can be interpreted as the Dirac operator associated with the cell complex~\cite{calmon2022higher,bianconi2021topological}.

When should we expect cell complexes to be more natural representations compared to hypergraphs? 
While we may reinterpret a cell complex as a specific type of hypergraph (see, e.g., the discussion in~\cite{torres2021and}), such a reinterpretation does in fact destroy the hierarchical notion of the relations contained in the complex, as any relations are simply treated as hyperedges, and the partial ordering between the respective cells remains unaccounted for.
This viewpoint if thus often not desired if the system of interest stems from a generalized geometric or physical context, and we are interested in \emph{different} types of (physical) data defined on nodes, edges, triangles, etc.
To provide some intuition from classical physics, note that the theorems of Stoke, Green, Gauss etc., all describe the relation of one physical quantity on the boundary to another one in the interior; these are essentially topological theorems~\cite{grady2010discrete} with a discrete equivalent expressible in terms of cell complexes.
Essentially, these relations describe (physics-enforced) compatibility conditions that need to hold at a boundary.

In cellular complexes, this type of coupling to a boundary is only mediated via cells of adjacent dimension (the boundary of a boundary is zero; intuitively, an object of dimension $k-2$ has zero measure for integration in a space of dimension $k$).
This does not imply that multi-body interactions between nodes cannot be retrieved from the structural description of a cell complex.
However, arguably, this aspect is not at the core of the construction of cellular complexes, but rather a property that becomes apparent when the complex is reinterpreted as a hypergraph, which however neglects the hierarchical properties of the complex structure.

\section{Combinatorial complexes: combining set-type and hierarchical interior-to-boundary relations}

\subsection{Hierarchical structure and set-type relations}
\label{gap}

As we have discussed above, in the study of higher-order networks, two essential features define the nature of relations: hierarchical structure, as encapsulated in cell complexes, and set-type relations \cite{hajijtopological}, as inherent in general hypergraphs.

\begin{definition}[Set-type relation]
A relation in a higher-order domain is called a \textit{set-type relation} if its existence is not implied by another relation in the domain.
\end{definition}

Set-type relations can be independently defined without being restricted by other relations. 

These features, hierarchical structure and set-type relations, are crucial in understanding and working with general higher-order networks. 
To allow the concurrent presence of these two features, we next recall the definitions of \emph{rank function} and \emph{combinatorial complex} \cite{hajijtopological}.

\begin{definition}[Rank function]\label{def:rank}
A \textit{rank function} on a higher-order domain $\mathcal{X}$ is an order-preserving function
$rk\colon \CCX\to \Znon$; i.e., $x\subseteq  y$ implies $rk(x) \leq rk(y)$ for all $x,y\in \CCX$. 
\end{definition}

A rank function assigns a (partial) order to the different relations in a higher-order domain. 
It allows us to organize relations in a hierarchy of interior-to-boundary relations, making it easier to understand the importance and complexity of each relation.
From this perspective, all complexes considered in \Cref{fig:data-structures} are naturally equipped with a canonical rank function. 
In particular,
a (hyper)graph admits the trivial rank function defined as $\rk(x)=0$ when $x$ is a vertex and $\rk(x)=1$ when $x$ is a (hyper)edge. 
A simplicial complex admits the rank function defined as $\rk(x) = |x|+1$ for $x \in \mathcal{X}$. Finally, a cell complex $\mathcal{X}$ admits the rank function $\rk(x_{\alpha})= n_{\alpha}$.

\subsection{Combinatorial complexes}

Combinatorial complexes (CCs) are higher-order topological domains
equipped with hierarchies of relations and set-type relations~\cite{hajijtopological}.
Thus, CCs constitute flexible domains for modeling and computing.
In addition, CCs amplify the space of solutions
to the structure learning problem
by providing a more diverse set of domains with enriched properties.

\begin{definition}[Combinatorial complex]
\label{def:cc}
A \textit{combinatorial complex (CC)} is a triple $(\mathcal V, \CCX, \rk)$ consisting of a set $\mathcal V$, a subset $\CCX$ of $\mathcal{P}(\mathcal V)\setminus\{\emptyset\}$, and a function
$\rk \colon \CCX\to \Znon$ with the following properties:
	 \begin{enumerate}
	 \item for all $v\in \mathcal V$, $\{v\}\in\CCX$ and $\rk(v)=0$,
	 \item the function $\rk$ is order-preserving,
  which means that if $x,y\in \CCX$ satisfy $x\subseteq  y$, then $\rk(x) \leq \rk(y)$.
	 \end{enumerate}
  Elements of $S$ are called \textit{entities} or \textit{vertices}, elements of $\CCX$ are called \textit{relations} or \textit{cells}, and $\rk$ is called the \textit{rank function} of the CC.
\end{definition}

\begin{figure}[tb!]
	\centering
 \includegraphics{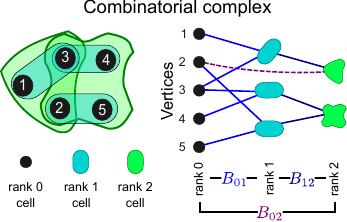}
	\caption{\textbf{Illustration of a combinatorial complex.} \textbf{Left}: Schematic drawing of a combinatorial complex. Black dots indicate rank-0 cells (vertices), cyan shapes indicate rank-1 cells, and light green indicates rank-2 cells. Note that vertex 2 is directly attached to the left rank-2 cell, without being part of a rank-1 cell that is subordinated to this rank-2 cell. Such a set-type hierarchical configuration is not possible for both cell complexes and hypergraphs.
    \textbf{Right}: a Hasse-like diagram of the multi-partite structure viewpoint of the combinatorial complex. Note that there are maps between cells of adjacency rank ($B_{01}$, $B_{12}$) similar to standard cell complexes. Furthermore, there is also a mapping $B_{02}$ from the rank-$2$ cells directly to rank-$0$ cells (in this case signified by the dashed violet link).
    The transposes of these maps naturally correspond to the (co-boundary) maps in the reverse direction.\vspace{-3mm}}
	\label{fig:comb_complex}
\end{figure}

The rank function $\rk$ of a CC stratifies subsets of relations in the CC and, therefore, builds hierarchies among such relations. 
CCs also permit set-type relations since there are no relation constraints in Definition~\ref{def:cc}. 
Thus, CCs subsume cell complexes and hypergraphs in the sense that they enable hierarchies of relations and set-type relations.
Figure~\ref{fig:comb_complex} displays an example of a CC.
The underlying multi-partite structure can be encoded via the following Dirac-like operator (for a complex of order three, as displayed in~\Cref{fig:comb_complex}):
\begin{equation}
    \mathbf{D}_\mathcal{CC} = 
\begin{bNiceMatrix}[first-row,first-col]
 & \mathcal V & \mathcal X_1 & \mathcal X_2 \\
\mathcal V & \mathbf{0} & \mathbf{B}_{01} & \mathbf{B}_{02} \\
\mathcal X_1 & \mathbf{B}_{01}^\top & \mathbf{0} & \mathbf{B}_{12} \\
\mathcal X_2 & \mathbf{B}_{02}^\top & \mathbf{B}_{12}^\top & \mathbf{0}
\end{bNiceMatrix} \in \mathbb{R}^{N_\mathcal{X}\times N_\mathcal{X}}.
\end{equation}
Note that while we can simply maintain the relation $\mathbf{B}_{01}\mathbf{B}_{12}=0$ analogous to cell complexes, defining a corresponding Hodge-decomposition in terms of the boundary operators $\mathbf{B}_{ij}$ that is aligned with the spectral properties of $\mathbf{D}_\mathcal{CC}$ is not possible in general (cf. ~\cite{grady2010discrete,schaub2021signal,lim2020hodge,mulas2022, hatcher2005algebraic}, for discussions on the Hodge-decomposition, its spectral characterization and utility for signal processing).


\subsection{An illustrative example: signal processing over combinatorial complexes}
We now provide a numerical experiment showcasing the advantage of learning over CCs.

\noindent\textbf{Dataset generation. }
Let $\mathcal{X}$ be a complex of order $3$. 
Without loss of generality, we assume that $\mathcal{X}$ is equipped with a rank function $\rk$. 
We associate to each cell of rank $k$ in the complex a scalar-valued signal, which we store in the vector $\mathbf{s_k} \in \mathbb{R}^{|\mathcal{X}_k|}$, i.e., $\mathbf{s_0}\in\mathbb{R}^{|\mathcal V|}$ is nothing but a graph signal.
We assemble all these different signals into a combined vector $\mathbf{s}={[\mathbf{s_0}^\top,\mathbf{s_1}^\top,\mathbf{s_2}^\top]}^\top$.
In a physical context, such signals may, e.g., correspond to scalar potentials associated to points (vertex signals), currents between those points (edge signals), and magnetic fluxes associated to areas (2-cells).

More specifically, we create a 2-dimensional simplicial complex $\mathcal{X}$ through Delaney triangulation applied to a set of $200$ randomly generated points on the plane,
resulting in a complex with $581$ edges and $382$ triangles.
We then assign the signals as follows. 
On the complex, we generate uniformly at random the flow signals $\mathbf{s}_1$ and $\mathbf{s}_2$. 
We then utilize these signals to define the vertex labels as follows.
Let us denote the compatibility vector of the triangles by $\mathbf{t}=\mathbf{s}_2 + \frac{1}{3} \mathbf{B}_{12}^T\mathbf{s}_1$.
For each triangle, the value in $\mathbf{t}$ is high if the sum of the (oriented) triangle signal and the average flow along an edge on the boundary of this triangle are aligned with the reference orientation.
The compatibility of a triangle is negative if these summed signals are against the reference orientation.
We set each vertex label according to the compatibility of the triangles in which the vertex participates.
Specifically, we set $\mathbf{s_0}_\text{,label}= u(\mathbf{B}_{02}\mathbf{t})$, where the matrix $\mathbf{B}_{02}$ is defined via the entries $[\mathbf{B}_{02}]_{ij} = 1$ if vertex $i$ is part of simplex $j$, and $0$ otherwise; and we use the step function $u(x)=1$ if $x>0$, and $u(x)=0$ otherwise.
So, a vertex label is $1$ if the sum of the compatibilities of all triangles it participates in is positive, and $0$ otherwise. Our task is to predict $\mathbf{s_0}_\text{,label}$, based on the observed initial signals $\mathbf{s}_1$ and $\mathbf{s}_2$.
The dataset is split into three subsets: $60\%$ for training, $20\%$ for validation, and $20\%$ for testing.


\noindent\textbf{Topological neural network setup. }
We perform label prediction by training two neural networks on $\mathcal{X}$, defined via
\begin{equation}
    \mathbf{s}_{\text{pred}}=\sigma_2 (\mathbf{P}\sigma_1 ( \mathbf{P} \mathbf{s} \mathbf{W}^{1} )\mathbf{W}^{2}),
\end{equation}
where $\sigma_1$ and $\sigma_2$ are set to be the ReLU and sigmoid activation functions, respectively.
$\mathbf{s}_{\text{pred}}$ is defined on the entire complex, whereas the label $\mathbf{s_0}_\text{,\,label}$ is defined only on the vertices. During training, only the portion of $\mathbf{s}_{\text{pred}}$ defined on vertices is compared to $\mathbf{s_0}_\text{,\,label}$.
The only difference between the two neural networks is the choice of shift operator $\mathbf{P}$.
One neural network leverages the structure of the simplicial complex $\mathcal{X}$ as is, by setting
$\mathbf{P}= \mathbf{I} + \mathbf{D}_\mathcal{X}$ \cite{roddenberry2021principled}.
The other neural network interprets $\mathcal{X}$ as a combinatorial complex $\mathcal{CC}$ and thus uses the shift operator
$\mathbf{P}= \mathbf{I} + \mathbf{D}_\mathcal{CC}$.
The neural networks are trained using cross-entropy loss and the Adam optimizer with a learning rate of $0.001$.

\noindent\textbf{Results. }
The accuracy (averaged over $3$ runs) of the predicted vertex labels is $0.90$ and $0.97$
for the neural networks based on the simplicial complex and combinatorial complex, respectively.
This result demonstrates that the additional structure encoded in a CC can improve predictions in applications in which there is a dependency between signals defined on $k$-cells with cardinality difference greater than one.
Intuitively, the reason is that even for simplical neural networks with `simplicial awareness'~\cite{roddenberry2021principled}, the transfer of information is more limited compared to message passing in CCs.



\vspace{-0.5mm}
\section*{Acknowledgment}
\footnotesize
Contribution Statement: All authors conceived the idea,  edited the paper and approved the final version. MTS wrote the first draft. MH conducted the numerical experiments.  

MH was supported in part by the National Science Foundation (NSF, DMS-2134231).
MTS acknowledges funding from the Ministry of Culture and Science (MKW) of the German State of North Rhine-Westphalia (NRW R\"uckkehrprogramm) and from the European Union (ERC, HIGH-HOPeS, 101039827). Views and opinions expressed are however those of MTS only and do not necessarily reflect those of the European Union or the European Research Council Executive Agency; neither the European Union nor the granting authority can be held responsible for them. 

\bibliographystyle{ieeetr}
\bibliography{references}

\end{document}